\documentclass[a4paper,conference]{IEEEtran}
\usepackage{times}
\usepackage{epsfig}
\usepackage{graphicx}
\usepackage{amsmath}
\usepackage{amssymb}
\usepackage{multirow}

\ifCLASSINFOpdf
\else
\fi
\begin{document}
%
\title{Multiple Future Prediction Leveraging Synthetic Trajectories}

\author{\IEEEauthorblockN{Lorenzo Berlincioni}
\IEEEauthorblockA{University of Florence \\
lorenzo.berlincioni@unifi.it}
\and
\IEEEauthorblockN{Federico Becattini}
\IEEEauthorblockA{University of Florence \\
federico.becattini@unifi.it}
\and
\IEEEauthorblockN{Lorenzo Seidenari}
\IEEEauthorblockA{University of Florence \\
lorenzo.seidenari@unifi.it}
\and
\IEEEauthorblockN{Alberto Del Bimbo}
\IEEEauthorblockA{University of Florence \\
alberto.delbimbo@unifi.it}}


%


\maketitle

\begin{abstract}
Trajectory prediction is an important task, especially in autonomous driving. The ability to forecast the position of other moving agents can yield to an effective planning, ensuring safety for the autonomous vehicle as well for the observed entities. In this work we propose a data driven approach based on Markov Chains to generate synthetic trajectories, which are useful for training a multiple future trajectory predictor. The advantages are twofold: on the one hand synthetic samples can be used to augment existing datasets and train more effective predictors; on the other hand, it allows to generate samples with multiple ground truths, corresponding to diverse equally likely outcomes of the observed trajectory. We define a trajectory prediction model and a loss that explicitly address the multimodality of the problem and we show that combining synthetic and real data leads to prediction improvements, obtaining state of the art results.
\end{abstract}


%
\IEEEpeerreviewmaketitle

\section{Introduction}

Trajectory prediction has become a major topic of research in computer vision for autonomous driving~\cite{lee2017desire, srikanth2019infer, alahi2016social, marchetti2020memnet, rhinehart2019precog, tang2019multiple, sadeghian2019sophie, choi2019drogon}. The task is of utmost importance, since predicting other agent trajectories permits to avoid danger and plan ego-motion safely.
Unfortunately, the autonomous driving datasets required to train prediction models are extremely expensive to gather effectively. Costly data acquisition campaigns are required to obtain large scale vehicle trajectories with context  and several sensors are needed: cameras, stereo pairs, LiDARs, IMUs and GPS.
Once the campaign is terminated, vehicle trajectories may be estimated by tracking detections and fusing LiDAR measurements \cite{Geiger2012CVPR}. Context is usually provided by remapping image semantic labels~\cite{chen2018deeplab} onto the ground plane, which also requires a LiDAR scan or a depth map. Finally ego-motion estimation is required to register multiple map and trajectory acquisitions over time.


Some dataset such as KITTI~\cite{Geiger2012CVPR}, TrafficPredict \cite{ma2019trafficpredict} or Argoverse \cite{chang2019argoverse} are acquired with instrumented cars using LiDAR and multiple cameras. Others are extracted using a multicamera setup, like NGISM \cite{ngsim} which has been collected at an US highway junction. Other than being costly, all these setups for data acquisition are extremely time consuming, requiring either to wait for data collection~\cite{ngsim} or to drive a car in the traffic for hours~\cite{Geiger2012CVPR, ma2019trafficpredict, chang2019argoverse}. This complexity has the effect of limiting the scale of the datasets.

Alternative ways to gather trajectory data is to rely on less expensive existing videos lacking sensor annotations and trying to estimate vehicle motion, for instance using SLAM~\cite{murORB2} or replacing sensor data with deep learning methods~\cite{becattini2019vehicle}. These methods however still require high quality videos captured from a moving vehicle.

To overcome data acquisition limitations, the use of synthetic datasets has always attracted the interest of deep learning researchers. The potential of using simulated data is the ability to increase the training data at little or no cost, thus making learned models more capable and robust. For instance, GANs have been used to generate synthetic eye imagery to train gaze estimators~\cite{shrivastava2017learning}. Synthetic images have also been used to train detectors in an automotive scenario~\cite{huang2018auggan}.

A different take on the problem is to generate completely synthetic data~\cite{saleh2018effective, richter2016playing, Dosovitskiy17}, using advanced game engines. What makes this so compelling is the possibility of controlling the rendering pipeline, which makes it possible to obtain pixel level annotations automatically at no cost.

Currently, there is no work addressing the training of trajectory predictors from synthetic data. Differently from images, trajectories are low dimensional, and are in principle easier to generate. Nonetheless, generated trajectories must be framed into a context in order to exploit knowledge about the surrounding environment at inference time. Moreover, trajectory data must be coherent in terms of scale and object dynamics.

In this paper we propose a procedural strategy for generating realistic synthetic pairs of trajectories and semantically labeled top-view maps, relying on statistics of existing datasets.
Computer graphics researches have often sought procedural methods to generate data~\cite{smelik2014survey}, which does not require costly handcrafting of digital artifacts by visual artists. In the specific case of city maps generation, recent methodologies combine terrain and water data to shape the city map~\cite{chen2008interactive, parish2001procedural}.
While these methods enable realistic designs of cities, our goal is slightly different. First, we do not need a whole city to be generated at once, since our prediction model has only access to a limited surrounding. This is in line with a feasible real-world system whose perception is limited by sensor range. It could be argued that the whole city could be generated and cached and then local snapshots could be retrieved. Nonetheless, our methodology allows to create a wider range of possibilities at a faster rate. The efficiency of our model coupled with its random nature makes it suitable for a deep learning training loop. Indeed, we are able to provide newly generated examples at learning time making the training set virtually infinite.

The main idea of this work is that roads are born from agents paths. While modern roads are designed from the need to connect locations and to optimize commerce and transport in general, some believe that in certain cases roads originated from men following trails drawn by animals~\cite{helbing2001self} such as the Icknield Way~\cite{icknield}. Relying on this principle it is easier to generate plausible trajectories and build maps around them, rather than generating a map or a city and fitting a plausible motion on it.

In addition, data acquired from sensors might not have access to all desired informations, which could instead be acquired in a synthetic or simulated environment. An example of this is occlusion caused by other vehicles, which has been addressed using GANs to generate samples recovering the structure of the layout~\cite{berlincioni2019road, bescos2019empty}. In the case of trajectory data, what can be observed in the real world is only the path taken by a vehicle. Yet, willing to predict its future location, multiple equally likely outcomes might be possible. This information is impossible to capture with sensors, while with synthetically generated data it is possible to offer a rose of possibilities for a single observation.

Overall, in this work we study the possibility of augmenting trajectory prediction datasets by generating synthetic data using a Markov Chain with parameters estimated from real data statistics. Our method consistently generates plausible trajectories paired with semantic context maps. Each sample is split into an observed past and a set of possible futures, meaning the observed variable and the variables to be predicted.

We show that our synthetic data can help in learning good features and that combined with real data can yield to state of the art results on trajectory prediction benchmarks.
The main contributions of this paper are:
\begin{itemize}
	\item We propose a method for estimating a Markov Chain describing vehicle dynamics from real data. This is then used to generate synthetic data to augment trajectory prediction datasets.
	\item Our generation pipeline allows us to create samples which explicitly address the multimodality of trajectory prediction, i.e. samples with a single past trajectory and multiple future outcomes that cover different roads.
	\item We propose a prediction model equipped with a recurrent controller that performs an incremental attention over possible future locations. By combining real and synthetic data we demonstrate that our model is able to achieve state of the art results.
	\item We introduce the novel \textit{Multimodality Loss}, which thanks to the generated multimodal samples, allows us to train the network with direct supervision on each possible future.
\end{itemize}

\newcommand{\imgwidth}{0.15\textwidth}

\begin{figure*}[t]
	\centering
	\includegraphics[width=\imgwidth]{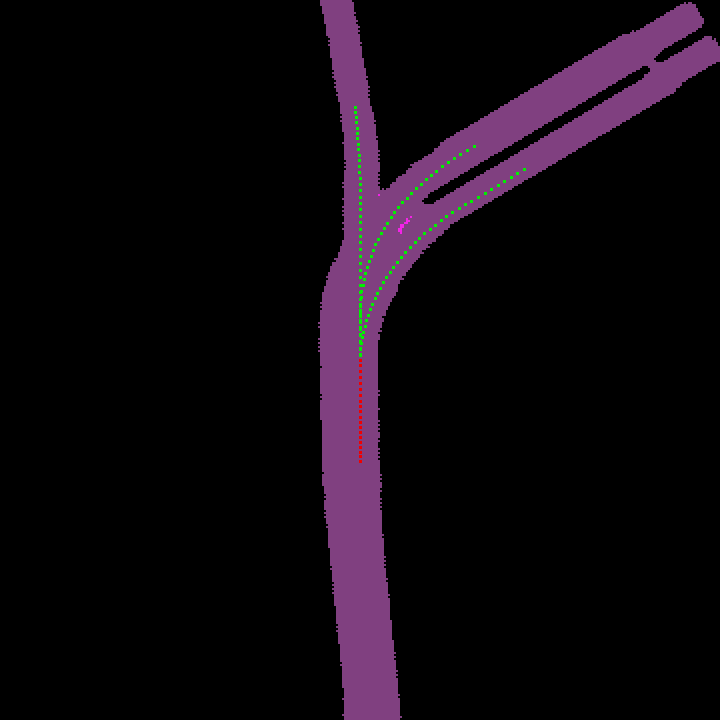}
	\includegraphics[width=\imgwidth]{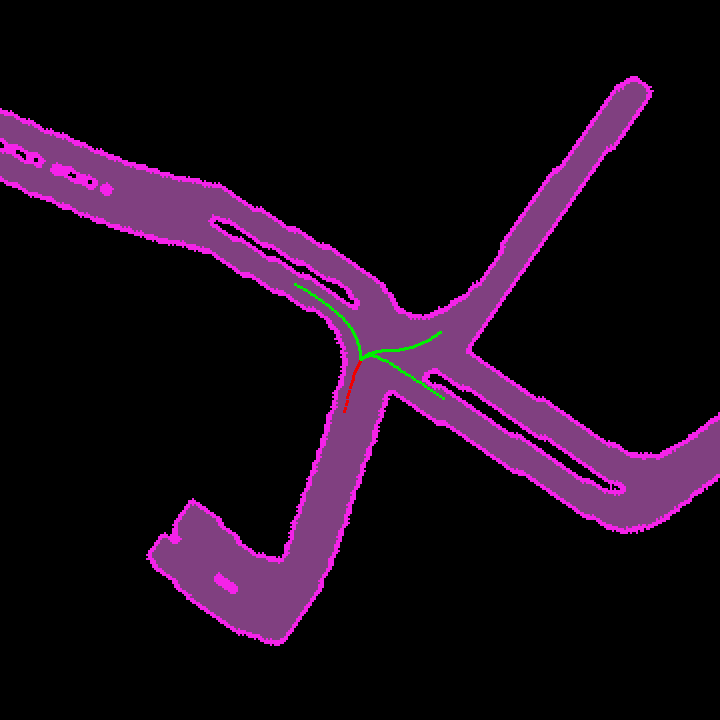}
	\includegraphics[width=\imgwidth]{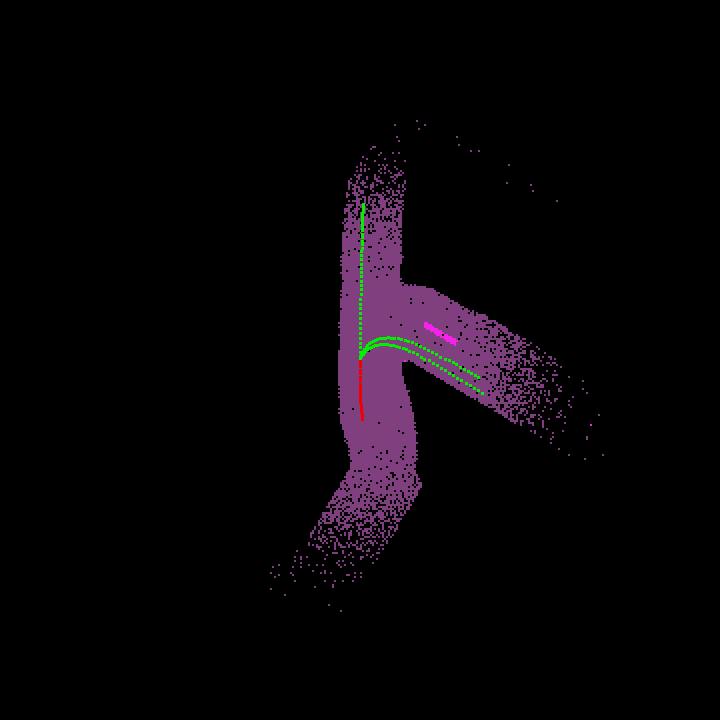}
	\includegraphics[width=\imgwidth]{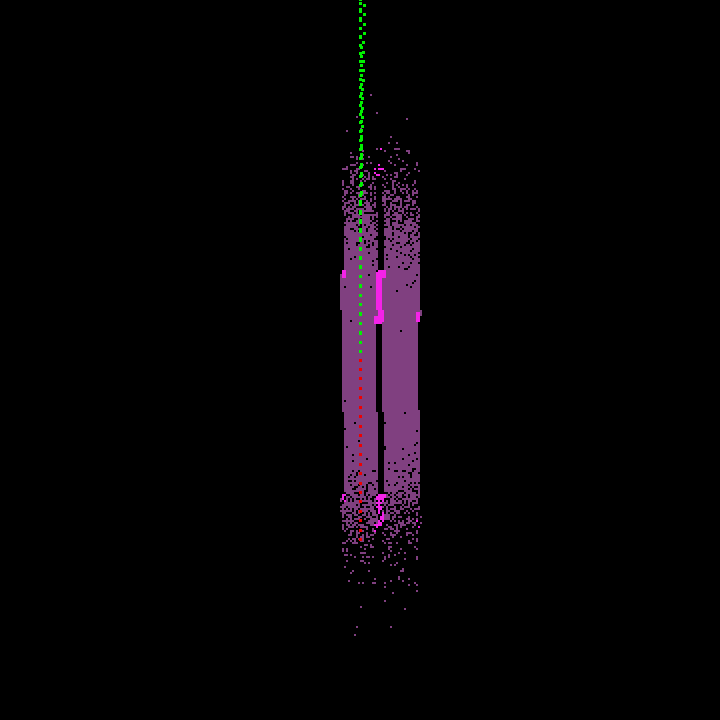}
	\includegraphics[width=\imgwidth]{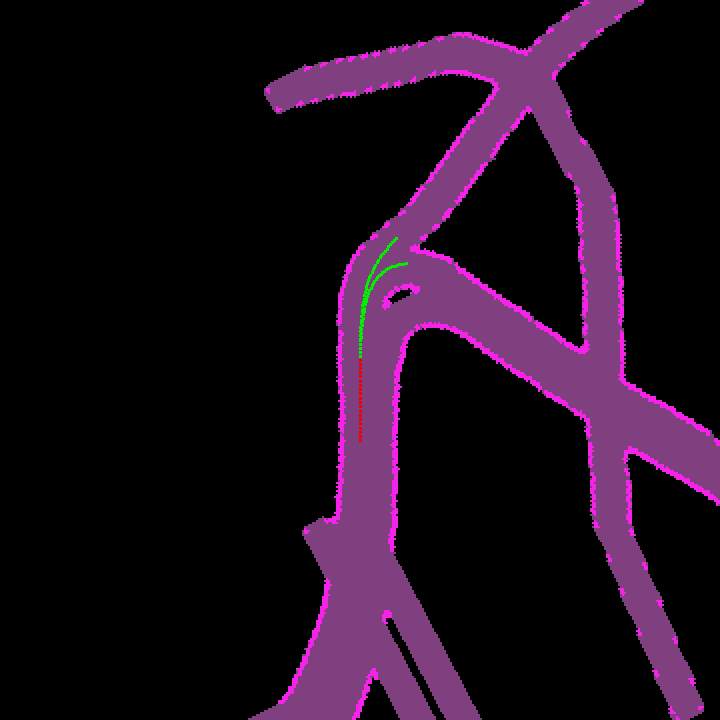}
	\includegraphics[width=\imgwidth]{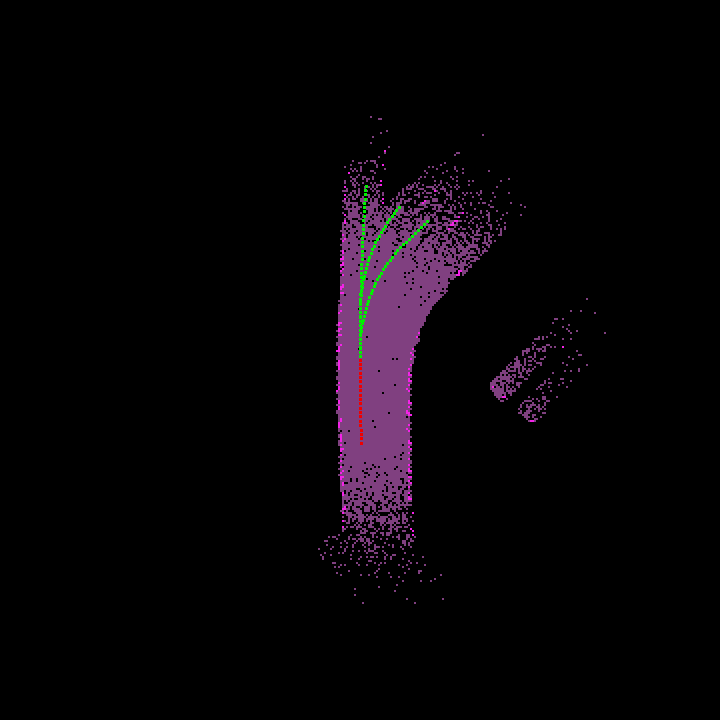} \\ \medskip
	\includegraphics[width=\imgwidth]{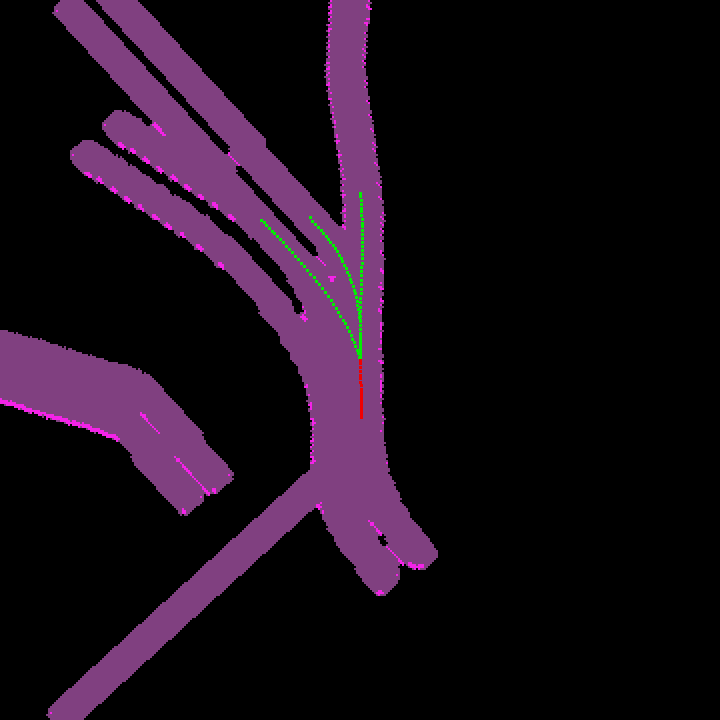}
	\includegraphics[width=\imgwidth]{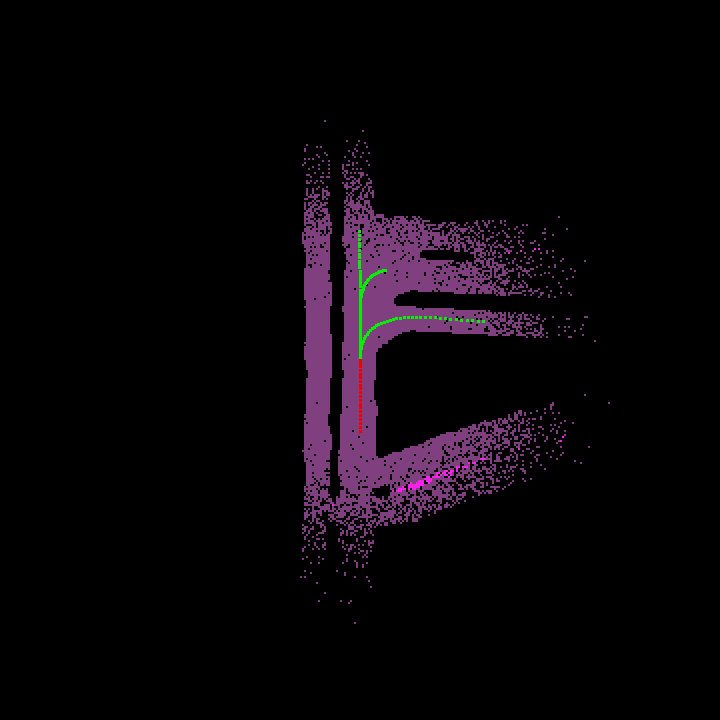}
	\includegraphics[width=\imgwidth]{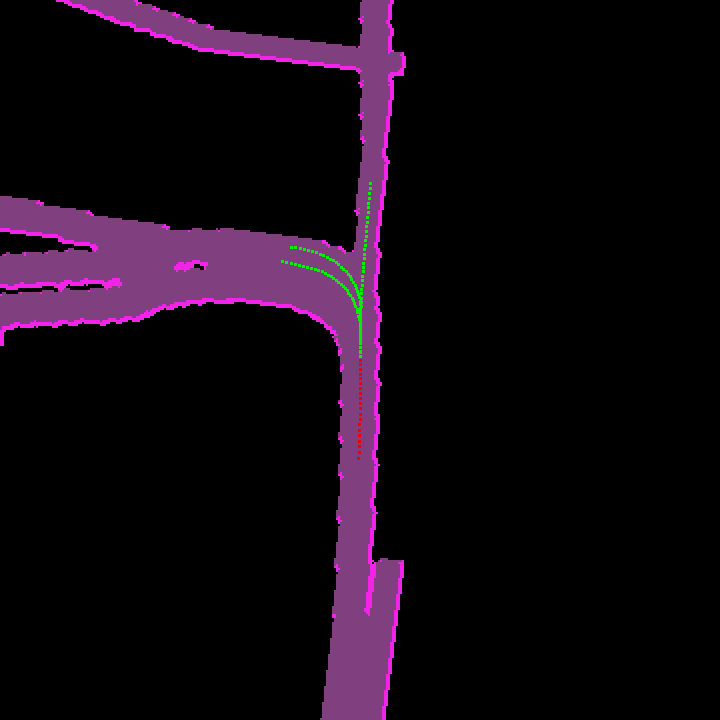}
	\includegraphics[width=\imgwidth]{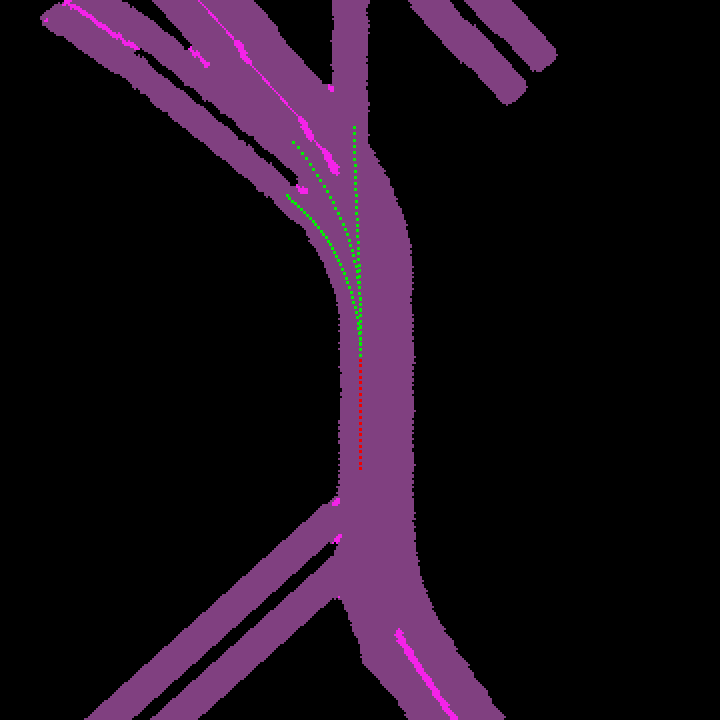}
	\includegraphics[width=\imgwidth]{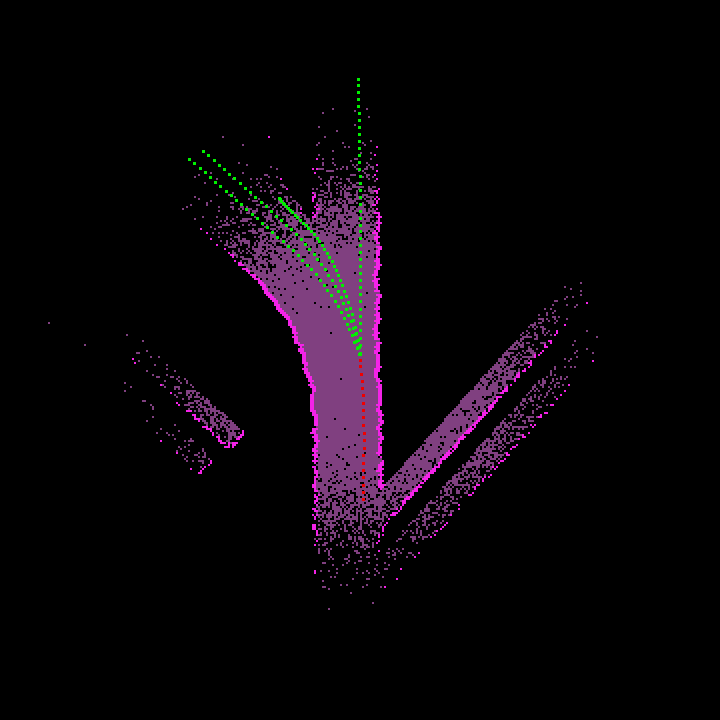}
	\includegraphics[width=\imgwidth]{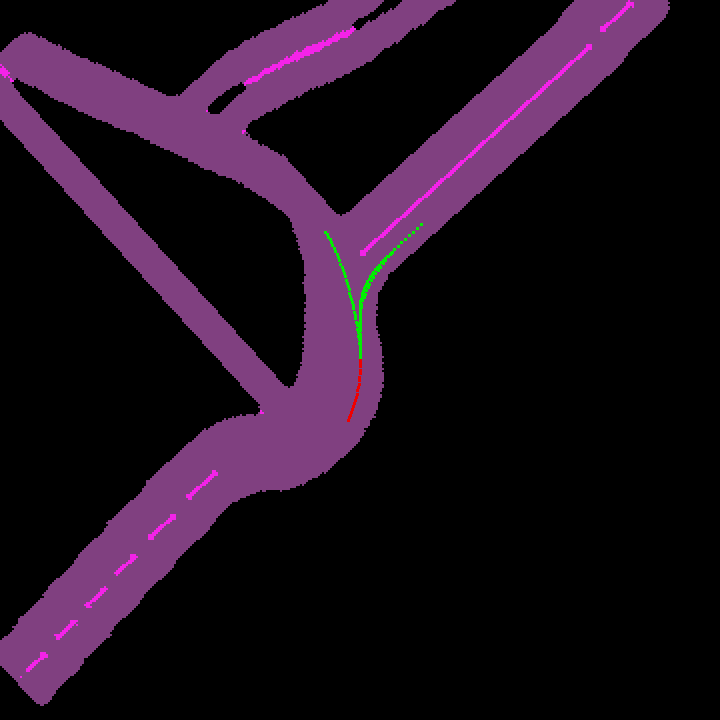}
	\caption{Synthetic trajectories in context. Semantic maps are $360\times360$px where a pixel corresponds to 0.5 meters. Purple corresponds to road pixels, pink to sidewalk and black to background. Trajectories are divided into 2s past (red) and multiple 4s futures (green) and are sampled at 10Hz.}
	\label{img:synthetic_maps}
\end{figure*}

\section{Synthetic Trajectory Generation}
We address the problem of vehicle trajectory prediction from a data-driven point of view. The focus of this work is to augment existing trajectory datasets with synthetic samples which can then be used to train predictive models more effectively.
Samples we want to generate are made of two main components: the actual trajectory followed by a vehicle and the context in which it is driving. We identify a trajectory as a sequence of coordinates $(x_i, y_i)$, each divided into two subsets: past coordinates and future coordinates. Past coordinates represent the observed history that a predictive model can observe, i.e. all positions occupied by the vehicle up to a given time identifiable as present, and future coordinates instead represent where the vehicle will go in the near future.
A context instead corresponds to a semantic map $c$, where each pixel is labeled with a category such as road or sidewalk.

To generate synthetic samples we model roads as paths carved by agents. Based on this idea, we use a random trajectory generator to draw paths and then we create in-scale semantic maps of roads and sidewalks.
Our goal is to have a fast method to generate a high variety of maps and trajectories on the fly rather than obtaining a full map of an urban or suburban scenario.
In the following we first outline our trajectory generation pipeline and then explain how to build complementary semantic maps.

\subsection{Trajectory Generation}
\label{sec:trajgen}
In order to generate synthetic trajectories we exploit a Markov Chain whose parameters are estimated from real data. The states of the chain correspond to vehicle position offsets from one timestep to the next. We represent offsets in a polar coordinate system with the y axis oriented in the same direction as the vehicle, therefore each state encodes speed and curvature of the vehicle in a given instant.

Given an initial random state, the Markov Chain allows to generate a trajectory through subsequent random state transitions. By concatenating all generated offsets we are then able to sample a complete trajectory, making a transition at each timestep. For example, if trajectories are sampled at 10Hz, then after  10 state transitions of the Markov Chain, a trajectory spanning over one second will be generated.

To identify the states of the Markov Chain and estimate the transition matrix, we rely on real trajectory data. Given a set of real trajectories in world coordinates $t_i = (x_0, y_0), (x_1, y_1), ..., (x_T, y_T)$ across $T+1$ timesteps, we first compute for each sample the T intermediate offsets in polar coordinates.
The radius $\rho_k$ is simply computed as the Euclidean distance between point $k+1$ and $k$ while the angle $\theta_k$ is equal to the change in orientation of the vehicle, i.e. the difference in degrees between the vehicle heading direction at time $k$ and $k+1$. This representation has the advantage of being rotation invariant, since the angles are computed relatively to the forward direction.

To obtain a finite and compact set of states, we apply K-means to cluster all offsets. The centroids of the discovered clusters represent an approximation of any possible state in which the vehicle can find itself, based on the training data. We use these centroids as nodes of the Markov Chain.

To estimate transitions we find all pairs of subsequent states in the dataset. Theoretically, each state could transition to any other, yet some transitions are physically implausible, i.e. if they imply sudden changes in speed or in steering angles. Each transition from a source state to a destination state is associated with a probability by counting their number of occurrences, normalized by the total number of transitions outgoing from the source node.
Such states and transitions define the Markov Chain we use for sampling new trajectories. A synthetic trajectory is built as a sequence of offsets belonging to the clusters encountered while visiting the Markov Chain. At each node a sample from the correspondent cluster is drawn and used to generate the current trajectory offset.

This process can be generalized to states that take into account multiple timesteps. In fact, by representing each state with a single cluster, each transition has a limited memory of the past evolution of the trajectory, which may result in erratic patterns.
To increase the memory, we simply identify each node in the chain with a sequence of temporally adjacent offsets, each quantized as a cluster centroid. In this way, a transition can be defined as a mapping from a sequence of N displacements occurred at timesteps (-N+1,..., -1, 0) to a sequence of displacements occurred at (-N+2,..., 0, 1), where timestep 0 corresponds to the present. Increasing N though will make the state space grow with a rate of $C^N$ where $C$ is the number of clusters, limiting at the same time the number of samples over which to estimate transition statistics. In our experiments we set N=2 unless differently stated.

\subsection{Map Generation}

A map $m$ is a tensor of size $H\times W \times C$ representing a top-view context labeled with semantic categories. $H$ and $W$ correspond to the spatial extent of the map and $C$ is the number of semantic classes used to label it. We use 3 classes, encoded as 1-hot vectors in each pixel of the map: road, sidewalk and background. We do not model other categories that can be typically found in urban scenes such as building or vegetation since they do not affect driving patterns. Each map has a granularity of $0.5$ meters per pixel, therefore a context covers an area of $H/2 \times W/2$ meters.

Since we are interested in modeling urban scenarios and vehicles move exclusively on roads, context maps are created by generating a set of trajectories and drawing roads around them. We use our trajectory generation pipeline to sample a sufficiently long path and, by adding a thick stroke to it, we are able to define the pixels labeled as road. In the same way, we add sidewalks next to lanes. The width of the stroke defines the width of a lane and its sidewalk. In our experiments we generate maps with lanes approximately 6 meters wide and sidewalks up to 1.5 meters, similarly to regular roads in the real world.

To obtain crossroads and forks we generate a new road starting from a random point along the previously generated one. We iterate this process a random number of times $b$, which we refer to as \textit{branching factor}. A higher branching factor leads to more complex scenes, while a branching factor of 1 provides a simple road with no intersections. In our experiments we use a branching factor up to 5.

To obtain richer scenarios, we randomly double the width of a whole generated path, indicating that the road has two lanes instead of one. Additional roads, either not connected to the main one or behind the vehicle, are added in the scene to include portions of the map that could potentially be taken by a vehicle, but that the vehicle we want to predict cannot reach. Despite this might seem unnecessary, we show that it helps the learning process of a predictive model, as discussed in Section~\ref{sec:ablation}.
Usually, maps similar to ours are obtained by combining LiDAR point clouds acquired the vehicle and semantic segmentation algorithms~\cite{Geiger2012CVPR}. This procedure though leads to noisy maps in regions far away from the sensor, since the point cloud gets sparser when the distance increases. To mimic this, we randomly add noise on map borders by turning road and sidewalk pixels to background.
Similarly, borders between categories tend to be noisy and irregular, therefore we randomly vary the width of sidewalks to simulate this effect.

\subsection{Multimodal synthetic sample generation}
To generate synthetic samples, comprising both a trajectory and its context, we select an $M$-point segment from one of the trajectories that generated the roads. The trajectory segment can then be split into two segments $p$ and $f$ of length $P$ and $F$ respectively, representing the past observation and the future trajectory.
The context is created by cropping a map centered in the present point, i.e. the last point of the past.
Throughout all experiments we set $P=20$ and $F=40$, which correspond to a 2 seconds past and a 4 seconds future with trajectories sampled at 10Hz, for a total of 6 seconds ($M=6$). The context instead is chosen to have an extent of $360\times360px$ ($180\times180m$).

To increase variability we shift trajectories orthogonally to the road by a random offset, with a higher probability to keep them close to the right side, as in an actual driving scenario assuming right-hand driving.

Generating synthetic samples has the immediate advantage of augmenting a trajectory prediction dataset. More importantly, simulated data can be generated to explicitly address the multimodality of the task. In fact, predicting the future position of a vehicle bares an intrinsic uncertainty, since multiple equally probably paths might be present, such as before intersections. Trajectory data collected from the real world cannot carry information about this multimodality, since a vehicle can only take a single direction out of the many possible ones.

Looking at the problem from a machine learning point of view, we want to learn a function that maps an observation $x$ into one of $K$ multiple outcomes $\{y_i\}_{i=1,...,K}$. In a supervised learning framework, real world data is able to provide a single supervision signal out of $K$. To make this worse, multiple examples might exist with similar observations $x$ and a completely different outcome $y$, which is detrimental to learning.
In a simulated environment instead, we can overcome this limitation by imagining several possible outcomes and providing all of them as ground truth to the learning algorithm.
To create multimodal trajectories, we simply select points in the future segment from which to initialize new trajectories and sample different transitions from the Markov Chain. By building roads around these trajectories, each encountered intersection will have an associated ground truth and each sample will have a set of possible outcomes.

Summarizing, a sample is made of: a semantic map $m$ centered in the present position of the vehicle; the past trajectory of the vehicle $p$; a set of $N_{GT}$ possible futures $f_i$ with $i=1,...,N_{GT}$.
Examples of synthetically generated maps with multiple futures are shown in Fig.~\ref{img:synthetic_maps}.

\begin{figure*}[htb]
	\centering
	\includegraphics[width=0.8\textwidth]{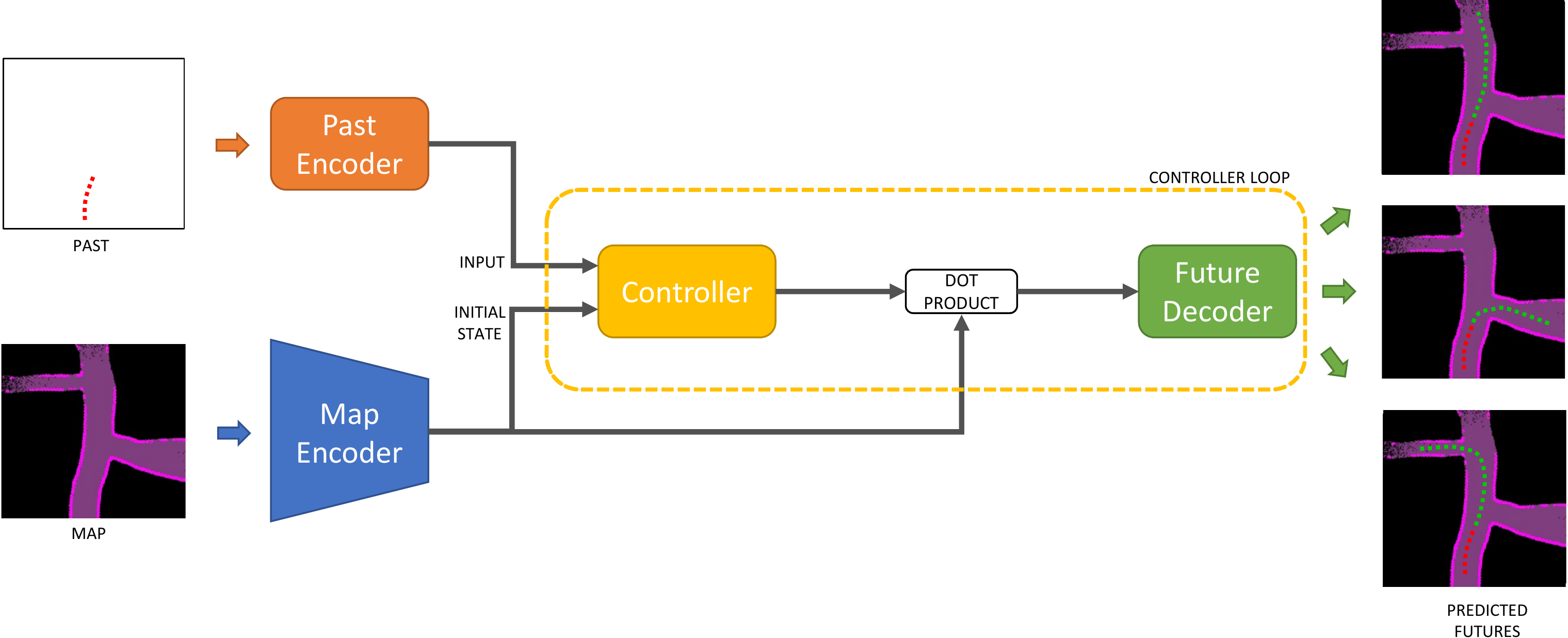}
	\caption{Architecture overview. Past trajectory and context map are encoded separately and used as input and initial state of the controller. The controller loops $K$ times and at each iteration performs an attention with the map encoding via dot product. The resulting vector is fed to the decoder which emits a prediction. A diverse future is obtained for each iteration of the controller.}
	\label{img:model}
\end{figure*}

\section{Prediction Model}
We developed a model specifically tailored to exploit synthetic samples with multimodal ground truth futures (Fig.~\ref{img:model}). The architecture is based on an encoder-decoder structure, which takes as input past trajectories and outputs multiple futures. Our model is equipped with a recurrent controller that at each step performs an attention on context maps, guiding the predictions towards different outcomes.
First, separate encoders learn latent representations for past and context. The trajectory encoder is a Gated Recurrent Unit (GRU) and the context encoder a Convolutional Neural Network (CNN). The two encoders are then fed to the controller, also implemented as a GRU. For each timestep, the same past is fed as input, while the context is used to initialize the hidden state. The memory of the GRU stores knowledge about future paths that have already been explored and outputs an attention vector which weighs the context embedding via dot product. The resulting vector is then fed to a final GRU that decodes it into a future prediction. This process is iterated $K$ times, where $K$ is the desired number of futures.

The recurrent layers, employed as encoder-decoder and controller, work with sequences on two different abstraction levels. The encoder and the decoder are modeling time, i.e. there is a correspondence between each update of the GRU and an actual timestep in the evolution of the vehicle dynamics. The controller, on the other hand, is modeling the multimodality of possible futures, exploring the semantic map to find possible roads that the vehicle might travel. At the same time, the controller is also modeling different modalities of navigating the same road (e.g. accelerating/decelerating).

\subsection{Implementation details}
All trajectories fed to the model, both at training and at testing time, are rotated such that the direction of the vehicle in the present follows an upward direction. This is useful since it provides rotation invariance and simplifies the taks.
The trajectory encoder network is implemented as a Recurrent Neural Network using a GRU with two layers with a hidden state size of 256.

\begin{figure}[htb]
	\centering
	\includegraphics[width=\columnwidth]{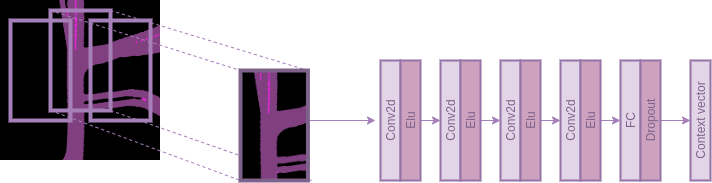}
	\caption{Architecture of the context map encoder. Multiple crops of the input map are extracted and encoded independently and successively combined.}
	\label{img:contextenc}
\end{figure}

The context encoder instead is a CNN network composed of 4 blocks of covolutional layers with ELu non linearities and a final fully connected layer, as shown in Fig. \ref{img:contextenc}. The context encoder receives multiple crops from the original top-view map and processes each one of them individually. We pick 3 overlapping crops in front of the position of the vehicle at time $t_0$ (the present), which coarsely represent the three main performable maneuvers (turn left, go straight, turn right). The advantage of doing so is to process the context map at a higher resolution without altering the structure of the network. The enconding vectors of each crop are finally concatenated and blended with a final fully connected layer to form a 256-dimensional representation.

The encodings for trajectory and context share the same dimension since we use them to initialize the controller, which is also implemented as a 256-dimensional GRU: the hidden state is initialized with the context and the trajectory is fed as input.

Finally, the trajectory decoder is a GRU with 3 layers and hidden state size of 256, followed by a fully connected layer that maps the output into the 2-dimensional offsets of the future predicted trajectory. The trajectory decoder is also trained with a dropout probability of 0.2.

\subsection{Training}
\label{sec:training}
The presence of the controller generating multiple futures, allows us to take full advantage of the synthetic trajectories, which are paired with several ground truths. In fact each ground truth can serve as supervision and each step of the GRU can be specifically optimized.
Usually, to enforce multiple diverse predictions, a \textit{Variety Loss}~\cite{gupta2018social} is used during training. This loss minimizes the Mean Squared Error between the only ground truth and the best prediction out of $K$ (this loss is sometimes referred to as \textit{best-of-K}). The advantage of doing backpropagation only through the best prediction is to avoid a single averaged solution and enforce the model to generate a set of diverse alternatives. This does not happen when optimizing the MSE of all generated futures compared with reference to a single ground truth.

Whereas this has often proven effective~\cite{gupta2018social, lee2017desire, srikanth2019infer, marchetti2020memnet}, it exploits only a partial supervision hence a large amount of computation during training is wasted not being used in backpropagation.
To overcome this limitation and exploit multiple synthetic ground truths, we introduce a \textit{Multimodality Loss} which optimizes a prediction for each available ground truth.
The loss computes pairwise distances between all targets and predictions. Then, it iteratively pairs the trajectories with the minimum distance in order to assign to each future at least a prediction. The first match is provided by the lowest pairwise distance. The paired GT and prediction are then temporarily removed and the process is repeated for all remaining ground truths. If $K>N_{GT}$, i.e. if the number of estimates is higher than the number of ground truths, the remaining predictions are paired to the closest future. In our experiments we use $K=5$ and a variable number of GT futures from 1 to 5.

The \textit{Multimodality Loss} allows us to backpropagate the error for each timestep of the controller, thus explicitly instructing the model about all possible future alternatives. We show in Section~\ref{sec:loss_ablation} that our loss provides benefits over existing losses such as MSE and \textit{Variety Loss}.

\begin{figure}[htb]
	\centering
	\includegraphics[width=\columnwidth]{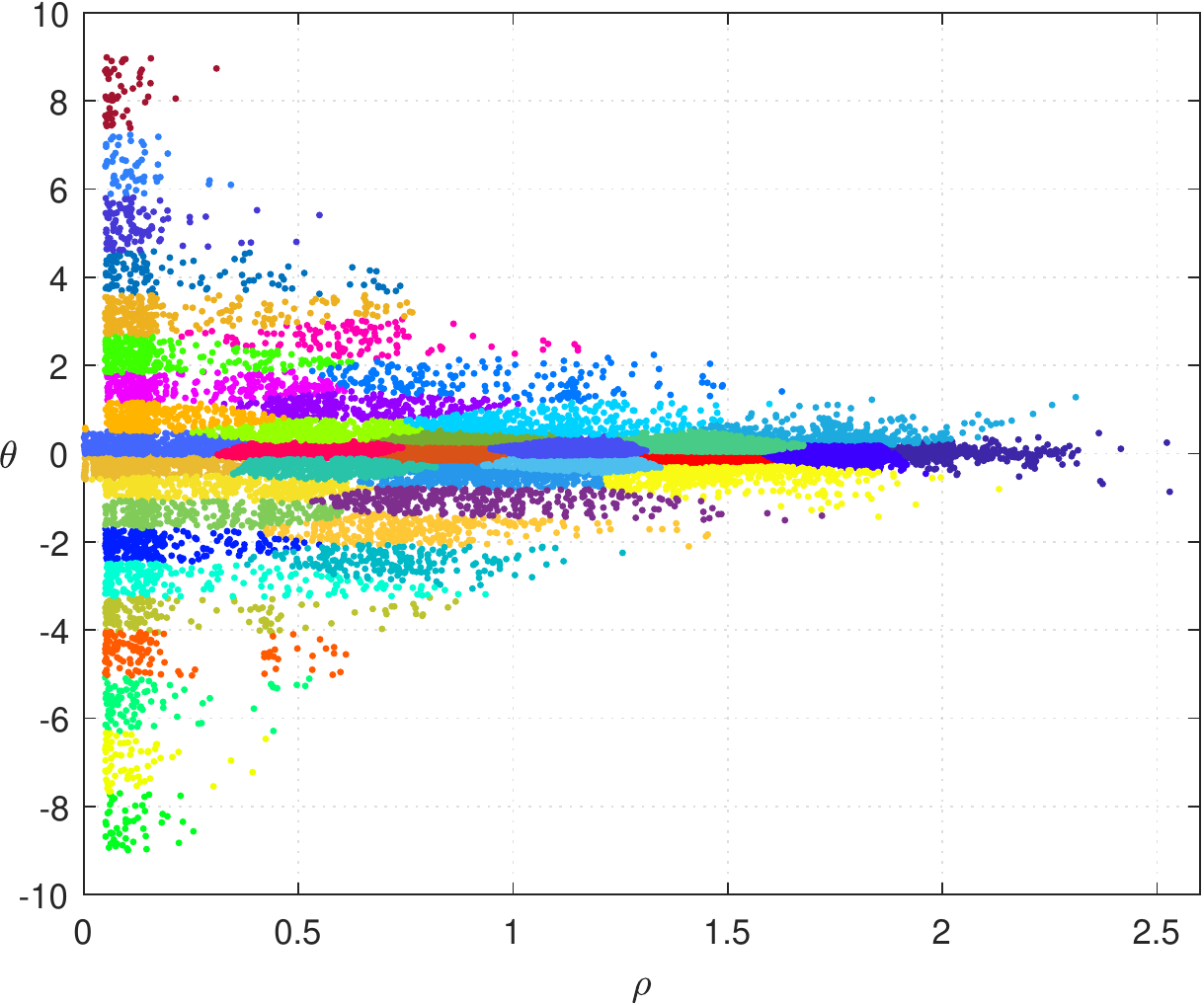}
	\caption{Trajectory offsets from the KITTI dataset in polar coordinates ($\rho, \theta$) clusterized through K-Means into 40 clusters. Different colors represent different clusters.}
	\label{img:clusters}
\end{figure}

\section{Experiments}
\subsection{Datasets and Metrics}
In our experiments we use the KITTI dataset~\cite{Geiger2012CVPR}, which comprehends several data modalities such as calibrated RGB streams, LiDAR 3D point clouds, annotated objects, semantic segmentations and IMU. Here we refer to the tracking dataset, which has often been used for trajectory prediction~\cite{lee2017desire, srikanth2019infer, marchetti2020memnet, marchetti2020multiple}. Despite this, several splits of the dataset have been used across prior work. Here we refer to the split introduced in~\cite{marchetti2020memnet}\footnote{https://github.com/Marchetz/KITTI-trajectory-prediction}, which contains 8613 top-view samples for training and 2907 for testing. Trajectories are divided into 2 seconds past trajectories and 4 seconds future trajectories, while maps have a spatial resolution of 0.5 meters per pixel. Trajectories are samples at 10Hz, therefore there are 20 and 40 points for past and future segments, respectively.

As metrics to test the performance of our model we measure the Final Displacement Error (FDE), i.e. the L2 error in meters at a given timestep (sometimes also referred to as Horizon Error), and Average Displacement Error (ADE), i.e. the error in meters averaged over all timesteps.
We compare our method against existing state of the art works~\cite{lee2017desire, srikanth2019infer, marchetti2020memnet} and some simpler baselines from~\cite{marchetti2020memnet}, namely a linear regressor, a multi layer perceptron regressor (MLP) and a Kalman Filter~\cite{kalman1960new}. It has to be noted that, due to the different dataset splits, \cite{lee2017desire} and \cite{srikanth2019infer} are not directly comparable to ours and are given as reference.

\subsection{Results}

To evaluate our model we first generate the states of the Markov Chain by applying K-means on trajectory offsets in polar coordinates extracted from the training set. Since trajectory coordinates in KITTI are acquired with GPS and IMU, they sometimes exhibit noise, especially when a vehicle is moving very slowly or not moving. To remove this noise we filter out offsets with $\rho<0.005$ and $\theta>0.5$ to prevent still vehicles to make sudden sharp turns. As discussed in Section~\ref{sec:ablation}, we found out the optimal number of clusters to be 40. Fig.~\ref{img:clusters} depicts the obtained clusters.

We trained three different variants of our method, varying the source of data: only real trajectories from KITTI, only synthetically generated trajectories, both real and synthetic trajectories. All variants are tested on the test set of KITTI, i.e. on real data.
Tab.~\ref{tab:results} shows the results obtained by the three methods, compared to prior work.
The usage of synthetic data alone is able to provide acceptable results: compared to its counterpart trained with real data, the model performs on par for predictions up to 2 seconds and with an FDE@4s only 0.5 meters worse. This is quite remarkable since no real sample is used to train the model, suggesting that our data generation process is able to approximate realistic samples. This result implies that sampling data using our Markov Chain could augment the existing dataset, thus improving the model without the need of costly data acquisition campaigns. In fact, this is the case when trained with mixed data. Here we use the training set from KITTI in combination with synthetic data. During training we sample approximately 16k synthetic samples, compared to the ~8k real ones, but we keep their ratio balanced in each batch. In this way, the error consistently lowers below the one obtained with real samples. Especially for far prediction horizons, the model is able to improve considerably, surpassing it by 0.7 meters of FDE@4s. In this way, we are able to improve also over existing prior work with the only exception of MANTRA~\cite{marchetti2020memnet} which is still better by a few centimeters at low time horizons. Samples of predicted trajectories are shown in Fig. \ref{img:preds_model}.

\begin{table}[]
	\caption{Average Displacement Error (ADE) and Final Displacement Error (FDE), computed for predictions at different time steps. DESIRE~\cite{lee2017desire} and INFER~\cite{srikanth2019infer} are shown as reference even if not directly comparable due to different dataset splits.}
	\label{tab:results}
	\resizebox{\columnwidth}{!}{
		
		\begin{tabular}{l|c|c|c|c||c|c|c|c}
			 & \multicolumn{4}{c||}{\textbf{ADE}} & \multicolumn{4}{c}{\textbf{FDE}} \\
			\textbf{Method} & 1s & 2s & 3s & 4s & 1s & 2s & 3s & 4s \\\hline
			
			Kalman~\cite{marchetti2020memnet}   &   0.51   &   1.14   &   1.99   &   3.03   &   0.97   &   2.54   &   4.71   &   7.41   \\
			Linear~\cite{marchetti2020memnet}   &   0.20   &   0.49   &   0.96   &   1.64   &   0.40   &   1.18   &   2.56   &   4.73   \\
			MLP~\cite{marchetti2020memnet}      &   0.20   &   0.49   &   0.93   &   1.53   &   0.40   &   1.17   &   2.39   &   4.12   \\
			MANTRA~\cite{marchetti2020memnet}   &   \textbf{0.17}   &   \textbf{0.36}   &   0.61   &   0.94   &   \textbf{0.30}   &   0.75   &   1.43   &   2.48   \\
			
			Ours (Synthetic data)   &  0.32    &  0.54    &  0.85    &  1.31    &  0.52    &  1.01    &  1.90    &  3.44    \\
			Ours (Real data)        &  0.31    &  0.53    &  0.78    &  1.24    &  0.51    &  0.95    &  1.63    &  2.95    \\
            Ours (Mixed data)       &  0.22    &  0.38    &  \textbf{0.59}    &  \textbf{0.89}    &  0.35    &  \textbf{0.73}    &  \textbf{1.29}    &  \textbf{2.27}    \\ \hline
			
			DESIRE~\cite{lee2017desire}         & - & - & - & - & 0.28 & 0.67 & 1.22 & 2.06 \\
			INFER~\cite{srikanth2019infer}      & 0.56 & 0.75 & 0.93 & 1.22 & 0.81 & 1.08 & 1.55 & 2.46 \\ \hline
			
	\end{tabular}}
\end{table}

\newcommand{\imgwidthpredcrops}{0.20\textwidth}
\begin{figure*}[htb]
	\centering
	\includegraphics[width=\imgwidthpredcrops]{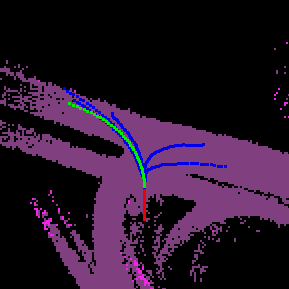}
	\includegraphics[width=\imgwidthpredcrops]{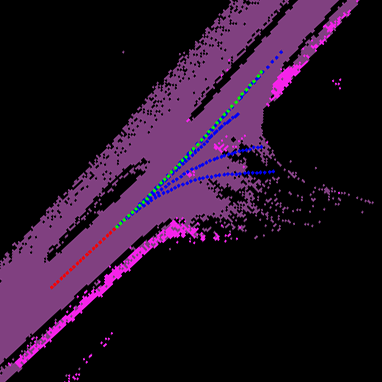}
	\includegraphics[width=\imgwidthpredcrops]{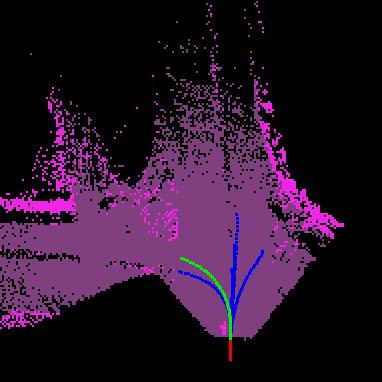}
	\includegraphics[width=\imgwidthpredcrops]{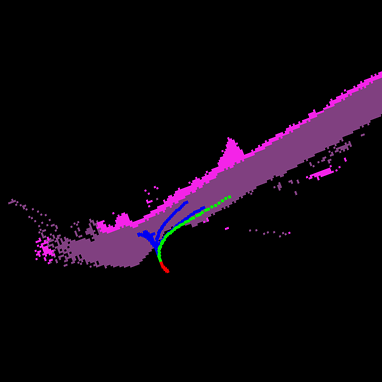} \\ \smallskip
	\includegraphics[width=\imgwidthpredcrops]{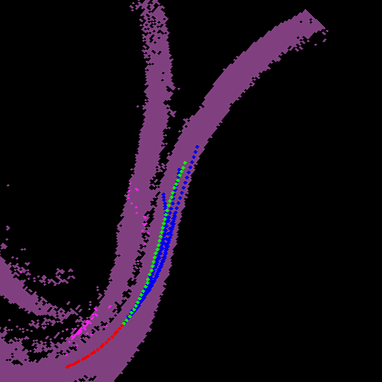}
	\includegraphics[width=\imgwidthpredcrops]{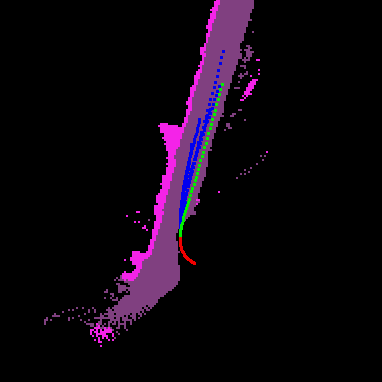}
	\includegraphics[width=\imgwidthpredcrops]{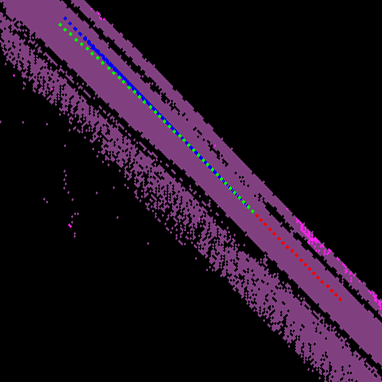}
	\includegraphics[width=\imgwidthpredcrops]{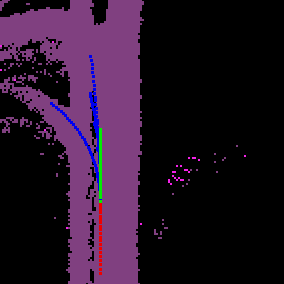}
	\caption{Outputs of our model trained with mixed data and tested on real data. Past trajectory in red, future trajectory in green and predictions in blue. Purple corresponds to road pixels, pink to sidewalk and black to background.}
	\label{img:preds_model}
\end{figure*}

\subsection{Effect of Multimodality Loss}
\label{sec:loss_ablation}
We investigated the advantage of using our \textit{Multimodality Loss} against standard losses such as \textit{Variety Loss} or simple MSE. As discussed in Sec.~\ref{sec:training}, the advantage of the \textit{Multimodality Loss} is to be able to optimize the network for each generated trajectory, instead of optimizing only the best prediction as the \textit{Variety Loss} would do. On the other hand, one could indeed backpropagate all predictions respect to a single ground truth, but this would lead to a lack of multimodality, generating averaged predictions that try to satisfy all possible likely futures.

In Tab.~\ref{tab:losses} we report the results obtained by the model using the three losses. For the \textit{Variety Loss} and the MSE we simply pick one of the possible ground truths and discard the information about the others during training. As expected, the MSE proves not the be suitable for the task as hand, due to its inability to generate diversity. As can be seen, our \textit{Multimodality loss} allows us to lower the error significantly even compared to the \textit{Variety Loss}, being able to effectively cover more future alternatives.

\begin{table}[]
	\caption{Analysis of the effect of different losses during training. Our \textit{Multimodality Loss} outperforms the \textit{Variety Loss} and MSE since it can explicitly address multimodality.}
	\label{tab:losses}
	\resizebox{\columnwidth}{!}{
		
		\begin{tabular}{l|c|c|c|c||c|c|c|c}
			& \multicolumn{4}{c||}{\textbf{ADE}} & \multicolumn{4}{c}{\textbf{FDE}} \\
			\textbf{Method} & 1s & 2s & 3s & 4s & 1s & 2s & 3s & 4s \\\hline
				
			MSE   &  0.35    &  0.68    &  1.16    &  1.81    &  0.59    &  1.42    &  2.75    &  4.68    \\
				
			\textit{Variety Loss}   &  0.34    &  0.54    &  0.80    &  1.19    &  0.53    &  0.94    &  1.70    &  3.03    \\
			
			\textit{Multimodality Loss}             &  \textbf{0.22}    &  \textbf{0.38}    &  \textbf{0.59}    &  \textbf{0.89}    &  \textbf{0.35}    &  \textbf{0.73}    &  \textbf{1.29}    &  \textbf{2.27}    \\ 
	\end{tabular}}
\end{table}

\subsection{Ablation Studies}
\label{sec:ablation}
We perform several ablation studies to analyze the importance of specific components in the model architecture and in the data generation process (Tab.~\ref{tab:ablation}).
First we trained our model disabling some components in the synthetic data generation process: without simulating LiDAR noise, without shifting trajectories inside lanes and without adding unreachable roads.

Turning off the synthetic LiDAR noise slightly lowers the performance of the model. This happens mostly due to vehicles with futures in noisy parts of the map. The model in fact interprets the noise as background and tries to avoid it.
Similar results are obtained when all generated trajectories are in the middle of the lane. Training with this data, the model often tends to make predictions drift towards the center of the road instead of following the natural path of the vehicle.
A more considerable drop in performance is observed without adding unreachable roads. When maps are generated with possible futures along every visible road, the controller tries to guide predictions towards both reachable and unreachable areas. This may lead to very unnatural predictions, since at test time the predicted paths will often cut through the background in order to reach every visible road.

We then tested the effect of using a different Markov Chain to generate trajectories. As explained in Sec.~\ref{sec:trajgen}, we normally use nodes that correspond to pairs of clusters, therefore taking two timesteps into account. We generated a Markov Chain with states composed of a single timestep and retrained the model. The generated samples do not approximate the real data well enough, leading to noisy trajectories that often change direction and speed abruptly. This reflects in a drop of 0.6 meters of FDE@4s as observed in Tab.~\ref{tab:ablation}.

What affects the model the most though is the attention mechanism. We trained our model disabling it, making the controller directly feed its output to the decoder. The map encoding is now taken into account only as initial state of the controller, instead of using it to guide individual predictions. This appears to be highly detrimental for the model, since the performance severely drop and the error rises by almost 3 meters at 4 seconds.

\begin{table}[]
	\caption{Ablation study. Our model is compared to variants with: no simulated LiDAR noise; no random trajectory shift across lanes; no unreachable roads; data generated by a Markov Chain with single timestep states; absence of controller.}
	\label{tab:ablation}
	\resizebox{\columnwidth}{!}{
		
		\begin{tabular}{l|c|c|c|c||c|c|c|c}
			& \multicolumn{4}{c||}{\textbf{ADE}} & \multicolumn{4}{c}{\textbf{FDE}} \\
			\textbf{Method} & 1s & 2s & 3s & 4s & 1s & 2s & 3s & 4s \\\hline
			
			Ours             &  \textbf{0.22}    &  \textbf{0.38}    &  \textbf{0.59}    &  \textbf{0.89}    &  \textbf{0.35}    &  \textbf{0.73}    &  \textbf{1.29}    &  \textbf{2.27}    \\ 
			
			No LiDAR noise   &  0.23    &  0.40    &  0.62    &  0.92    &  0.37    &  0.75    &  1.34    &  2.35    \\
			
			No trajectory shift     &  0.26    &  0.45    &  0.68    &  0.99    &  0.43    &  0.83    &  1.42    &  2.40    \\

			No unreachable roads   &  0.29    &  0.48    &  0.72    &  1.06    &  0.47    &  0.87    &  1.50    &  2.62    \\
			
			Single chain states   &  0.37    &  0.55    &  0.81    &  1.18    &  0.54    &  0.97    &  1.68    &  2.91    \\
									
			No attention     &  0.42    &  0.80    &  1.31    &  2.02    &  0.70    &  1.61    &  3.03    &  5.15    \\
			
	\end{tabular}}
\end{table}

\begin{figure}[htb]
	\centering
	\includegraphics[width=\columnwidth]{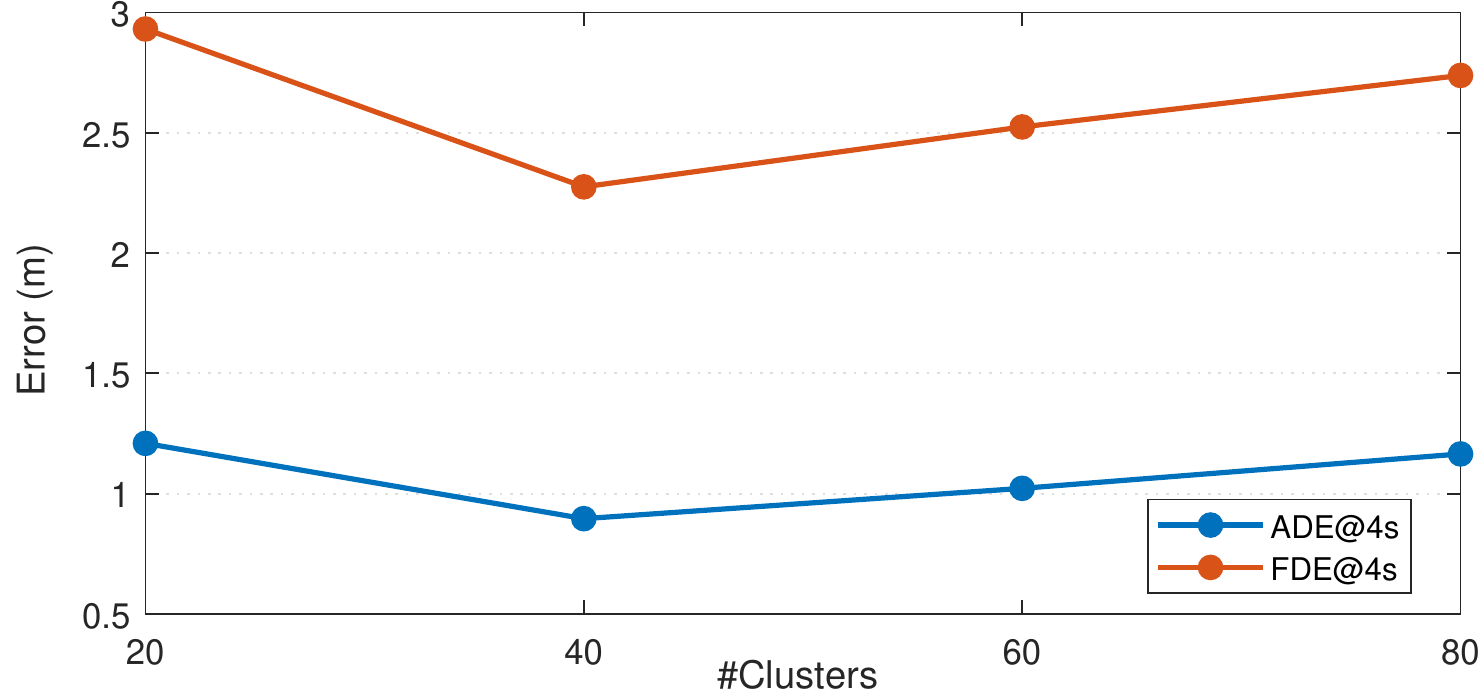}
	\caption{Results obtained varying the number of clusters in K-Means.}
	\label{img:num_clusters}
\end{figure}

In addition we verified the effect of the number of clusters for K-Means when generating the states for the Markov Chain. Fig.~\ref{img:num_clusters} shows the resulting FDE and ADE at a time horizon of 4 seconds using a number of clusters equal to 20, 40, 60 and 80. It appears that the optimal value is 40 and that the error curve is convex with reference to the number of clusters. The model however is quite robust to changes since the FDE remains under 3 meters for all tested values.

\begin{figure}[htb]
	\centering
	\includegraphics[width=0.32\columnwidth, trim={175px 250px 175px 100px},clip]{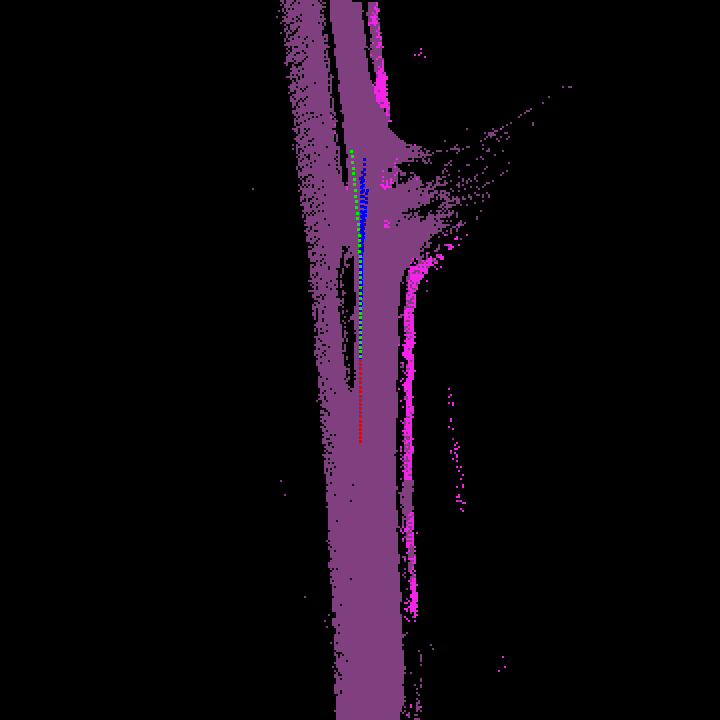}
	\includegraphics[width=0.32\columnwidth, trim={200px 250px 250px 200px},clip]{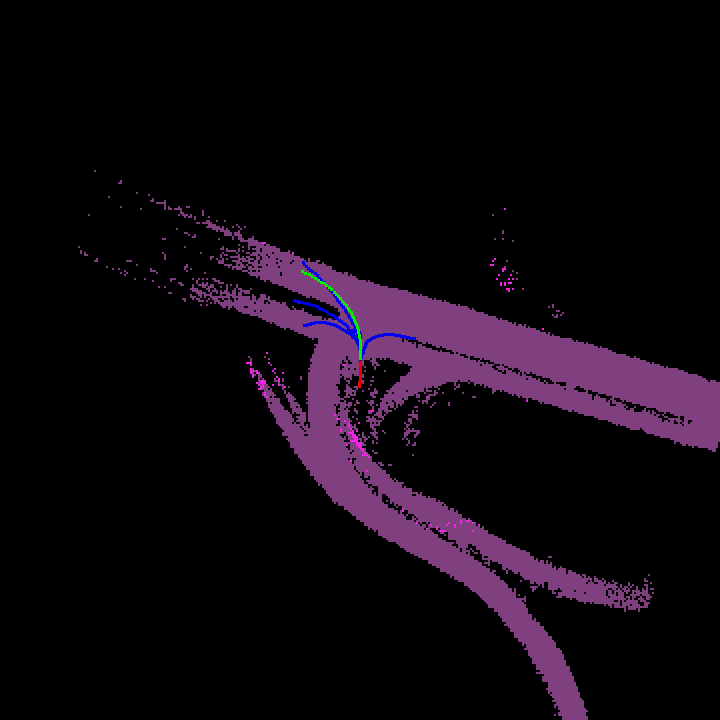} 
	\includegraphics[width=0.32\columnwidth, trim={75px 200px 275px 150px},clip]{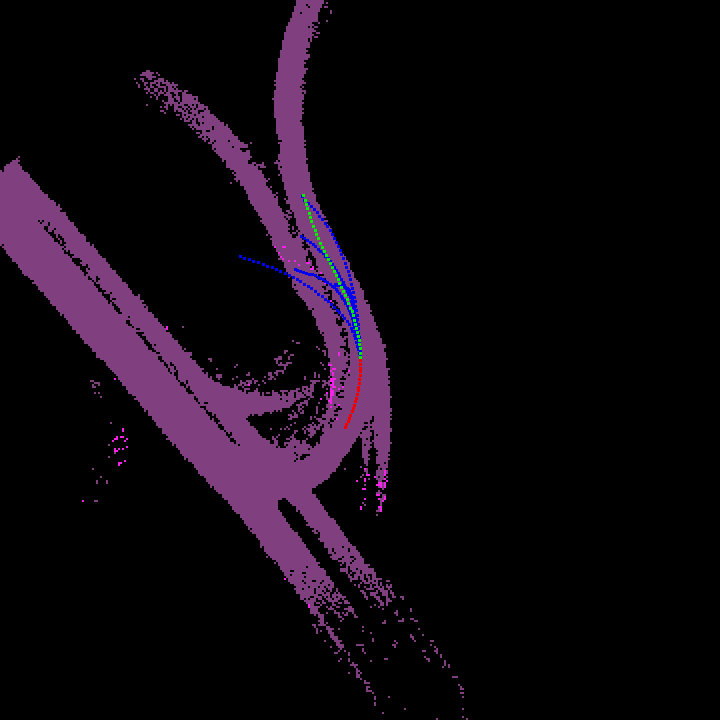}\\ \smallskip
	
	\includegraphics[width=0.32\columnwidth, trim={175px 250px 175px 100px},clip]{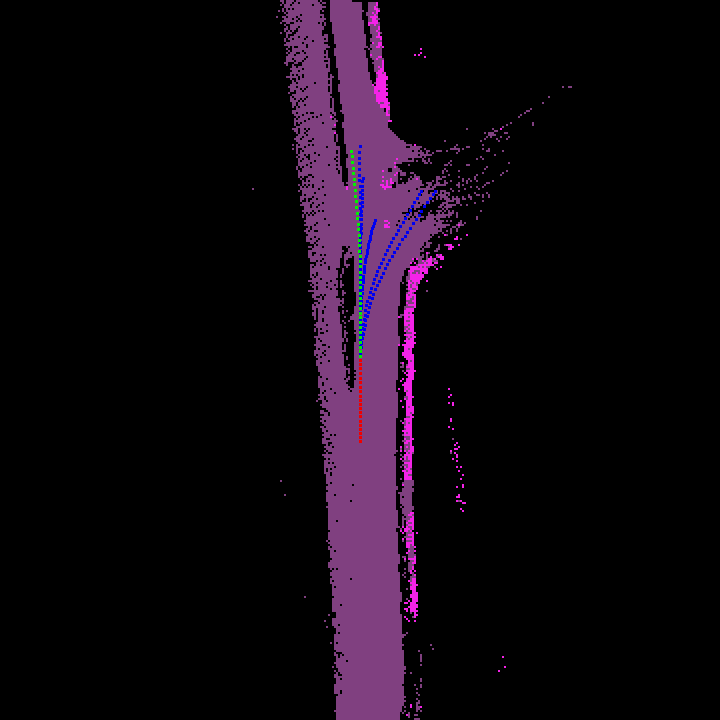}
	\includegraphics[width=0.32\columnwidth, trim={200px 250px 250px 200px},clip]{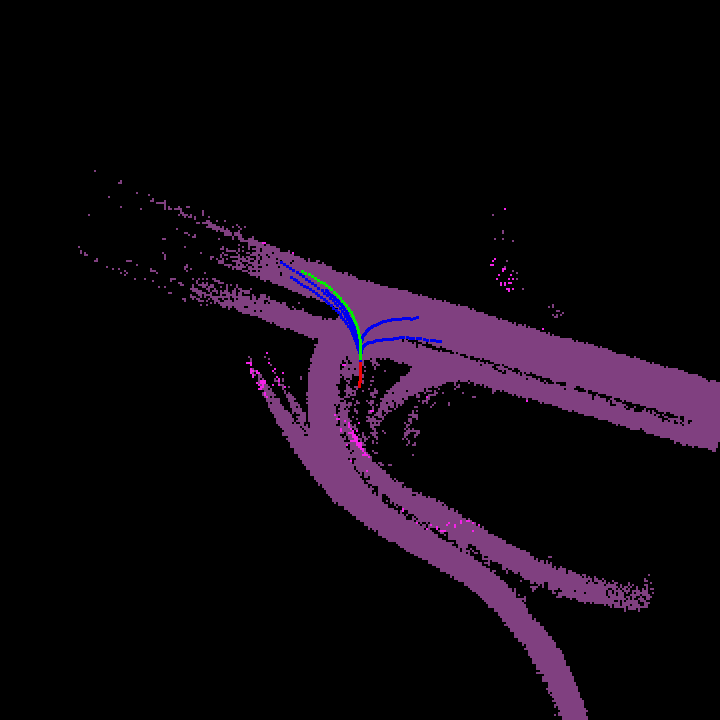}
	\includegraphics[width=0.32\columnwidth, trim={75px 200px 275px 150px},clip]{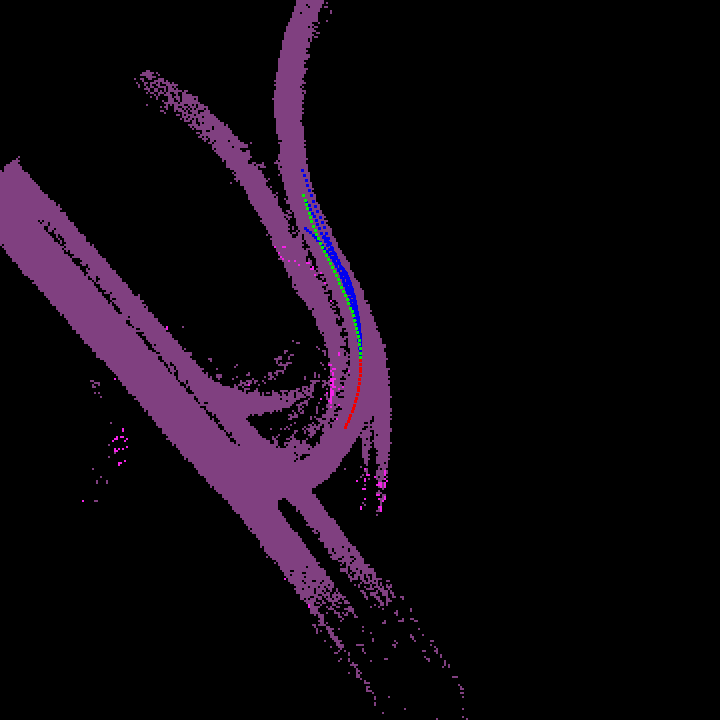}
	
	\caption{Ablations study samples on real data. Predictions obtained training the model without (top) and with (bottom) adopting synthetic data augmentation strategies: \textit{LiDAR noise} (left); \textit{Trajectory shift} (middle); \textit{Unreachable roads} (right). \vspace{-10px}}
	\label{img:ablation}
\end{figure}

\section{Conclusion}
In this paper we presented a method to generate synthetic trajectory samples exploiting a Markov Chain with parameters estimated from real data. This has shown two main advantages. First, the possibility to augment existing datasets and train better prediction models. Second, the possibility to couple past observations with multiple ground truths, which allowed us to exploit a new loss to train our model with full supervision and address the intrinsic multimodality of the task. The usage of this technique for generating synthetic data, along with a model specifically tailored for multimodal predictions, has led to state of the art results on the KITTI trajectory prediction benchmark.

\section*{Acknowledgments}
\footnotesize
This work was supported by the European Commission under European Horizon 2020 Programme, grant number 951911 - AI4Media.



%
%
%

\bibliographystyle{IEEEtran}
\bibliography{egbib}

\begin{thebibliography}{10}
\providecommand{\url}[1]{#1}
\csname url@samestyle\endcsname
\providecommand{\newblock}{\relax}
\providecommand{\bibinfo}[2]{#2}
\providecommand{\BIBentrySTDinterwordspacing}{\spaceskip=0pt\relax}
\providecommand{\BIBentryALTinterwordstretchfactor}{4}
\providecommand{\BIBentryALTinterwordspacing}{\spaceskip=\fontdimen2\font plus
\BIBentryALTinterwordstretchfactor\fontdimen3\font minus
  \fontdimen4\font\relax}
\providecommand{\BIBforeignlanguage}[2]{{%
\expandafter\ifx\csname l@#1\endcsname\relax
\typeout{** WARNING: IEEEtran.bst: No hyphenation pattern has been}%
\typeout{** loaded for the language `#1'. Using the pattern for}%
\typeout{** the default language instead.}%
\else
\language=\csname l@#1\endcsname
\fi
#2}}
\providecommand{\BIBdecl}{\relax}
\BIBdecl

\bibitem{lee2017desire}
N.~Lee, W.~Choi, P.~Vernaza, C.~B. Choy, P.~H. Torr, and M.~Chandraker,
  ``Desire: Distant future prediction in dynamic scenes with interacting
  agents,'' in \emph{Proceedings of the IEEE Conference on Computer Vision and
  Pattern Recognition}, 2017, pp. 336--345.

\bibitem{srikanth2019infer}
S.~Srikanth, J.~A. Ansari, S.~Sharma \emph{et~al.}, ``Infer: Intermediate
  representations for future prediction,'' \emph{arXiv preprint
  arXiv:1903.10641}, 2019.

\bibitem{alahi2016social}
A.~Alahi, K.~Goel, V.~Ramanathan, A.~Robicquet, L.~Fei-Fei, and S.~Savarese,
  ``Social lstm: Human trajectory prediction in crowded spaces,'' in
  \emph{Proceedings of the IEEE Conference on Computer Vision and Pattern
  Recognition}, 2016, pp. 961--971.

\bibitem{marchetti2020memnet}
F.~Marchetti, F.~Becattini, L.~Seidenari, and A.~Del~Bimbo, ``Mantra: Memory
  augmented networks for multiple trajectory prediction,'' in \emph{Proceedings
  of the IEEE Conference on Computer Vision and Pattern Recognition}, 2020.

\bibitem{rhinehart2019precog}
N.~Rhinehart, R.~McAllister, K.~Kitani, and S.~Levine, ``Precog: Prediction
  conditioned on goals in visual multi-agent settings,'' in \emph{Proceedings
  of the IEEE International Conference on Computer Vision}, 2019, pp.
  2821--2830.

\bibitem{tang2019multiple}
C.~Tang and R.~R. Salakhutdinov, ``Multiple futures prediction,'' in
  \emph{Advances in Neural Information Processing Systems}, 2019, pp.
  15\,398--15\,408.

\bibitem{sadeghian2019sophie}
A.~Sadeghian, V.~Kosaraju, A.~Sadeghian, N.~Hirose, H.~Rezatofighi, and
  S.~Savarese, ``Sophie: An attentive gan for predicting paths compliant to
  social and physical constraints,'' in \emph{Proceedings of the IEEE
  Conference on Computer Vision and Pattern Recognition}, 2019, pp. 1349--1358.

\bibitem{choi2019drogon}
C.~Choi, A.~Patil, and S.~Malla, ``Drogon: A causal reasoning framework for
  future trajectory forecast,'' \emph{arXiv preprint arXiv:1908.00024}, 2019.

\bibitem{Geiger2012CVPR}
A.~Geiger, P.~Lenz, and R.~Urtasun, ``Are we ready for autonomous driving? the
  kitti vision benchmark suite,'' in \emph{Conference on Computer Vision and
  Pattern Recognition (CVPR)}, 2012.

\bibitem{chen2018deeplab}
L.-C. Chen, G.~Papandreou, I.~Kokkinos, K.~Murphy, and A.~L. Yuille, ``Deeplab:
  Semantic image segmentation with deep convolutional nets, atrous convolution,
  and fully connected crfs,'' \emph{IEEE transactions on pattern analysis and
  machine intelligence}, vol.~40, no.~4, pp. 834--848, 2018.

\bibitem{ma2019trafficpredict}
Y.~Ma, X.~Zhu, S.~Zhang, R.~Yang, W.~Wang, and D.~Manocha, ``Trafficpredict:
  Trajectory prediction for heterogeneous traffic-agents,'' in
  \emph{Proceedings of the AAAI Conference on Artificial Intelligence},
  vol.~33, 2019, pp. 6120--6127.

\bibitem{chang2019argoverse}
M.-F. Chang, J.~Lambert, P.~Sangkloy, J.~Singh, S.~Bak, A.~Hartnett, D.~Wang,
  P.~Carr, S.~Lucey, D.~Ramanan \emph{et~al.}, ``Argoverse: 3d tracking and
  forecasting with rich maps,'' in \emph{Proceedings of the IEEE Conference on
  Computer Vision and Pattern Recognition}, 2019, pp. 8748--8757.

\bibitem{ngsim}
U.~F.~H. Administration, ``Us highway 101 dataset,'' 2005".

\bibitem{murORB2}
R.~Mur-Artal and J.~D. Tard\'os, ``{ORB-SLAM2}: an open-source {SLAM} system
  for monocular, stereo and {RGB-D} cameras,'' \emph{IEEE Transactions on
  Robotics}, vol.~33, no.~5, pp. 1255--1262, 2017.

\bibitem{becattini2019vehicle}
F.~Becattini, L.~Seidenari, L.~Berlincioni, L.~Galteri, and A.~Del~Bimbo,
  ``Vehicle trajectories from unlabeled data through iterative plane
  registration,'' in \emph{International Conference on Image Analysis and
  Processing}.\hskip 1em plus 0.5em minus 0.4em\relax Springer, 2019, pp.
  60--70.

\bibitem{shrivastava2017learning}
A.~Shrivastava, T.~Pfister, O.~Tuzel, J.~Susskind, W.~Wang, and R.~Webb,
  ``Learning from simulated and unsupervised images through adversarial
  training,'' in \emph{Proceedings of the IEEE conference on computer vision
  and pattern recognition}, 2017, pp. 2107--2116.

\bibitem{huang2018auggan}
S.-W. Huang, C.-T. Lin, S.-P. Chen, Y.-Y. Wu, P.-H. Hsu, and S.-H. Lai,
  ``Auggan: Cross domain adaptation with gan-based data augmentation,'' in
  \emph{Proceedings of the European Conference on Computer Vision (ECCV)},
  2018, pp. 718--731.

\bibitem{saleh2018effective}
F.~S. Saleh, M.~S. Aliakbarian, M.~Salzmann, L.~Petersson, and J.~M. Alvarez,
  ``Effective use of synthetic data for urban scene semantic segmentation,'' in
  \emph{European Conference on Computer Vision}.\hskip 1em plus 0.5em minus
  0.4em\relax Springer, 2018, pp. 86--103.

\bibitem{richter2016playing}
S.~R. Richter, V.~Vineet, S.~Roth, and V.~Koltun, ``Playing for data: Ground
  truth from computer games,'' in \emph{European Conference on Computer
  Vision}.\hskip 1em plus 0.5em minus 0.4em\relax Springer, 2016, pp. 102--118.

\bibitem{Dosovitskiy17}
A.~Dosovitskiy, G.~Ros, F.~Codevilla, A.~Lopez, and V.~Koltun, ``{CARLA}: {An}
  open urban driving simulator,'' in \emph{Proceedings of the 1st Annual
  Conference on Robot Learning}, 2017, pp. 1--16.

\bibitem{smelik2014survey}
R.~M. Smelik, T.~Tutenel, R.~Bidarra, and B.~Benes, ``A survey on procedural
  modelling for virtual worlds,'' in \emph{Computer Graphics Forum}, vol.~33,
  no.~6.\hskip 1em plus 0.5em minus 0.4em\relax Wiley Online Library, 2014, pp.
  31--50.

\bibitem{chen2008interactive}
G.~Chen, G.~Esch, P.~Wonka, P.~M{\"u}ller, and E.~Zhang, ``Interactive
  procedural street modeling,'' in \emph{ACM transactions on graphics (TOG)},
  vol.~27, no.~3.\hskip 1em plus 0.5em minus 0.4em\relax ACM, 2008, p. 103.

\bibitem{parish2001procedural}
Y.~I. Parish and P.~M{\"u}ller, ``Procedural modeling of cities,'' in
  \emph{Proceedings of the 28th annual conference on Computer graphics and
  interactive techniques}.\hskip 1em plus 0.5em minus 0.4em\relax ACM, 2001,
  pp. 301--308.

\bibitem{helbing2001self}
D.~Helbing, P.~Moln{\'a}r, I.~J. Farkas, and K.~Bolay, ``Self-organizing
  pedestrian movement,'' \emph{Environment and planning B: planning and
  design}, vol.~28, no.~3, pp. 361--383, 2001.

\bibitem{icknield}
\BIBentryALTinterwordspacing
I.~W. Association. (2019) Icknield way. [Online]. Available:
  \url{http://www.icknieldwaypath.co.uk/}
\BIBentrySTDinterwordspacing

\bibitem{berlincioni2019road}
L.~Berlincioni, F.~Becattini, L.~Galteri, L.~Seidenari, and A.~Del~Bimbo,
  ``Road layout understanding by generative adversarial inpainting,'' in
  \emph{Inpainting and Denoising Challenges}.\hskip 1em plus 0.5em minus
  0.4em\relax Springer, 2019, pp. 111--128.

\bibitem{bescos2019empty}
B.~Bescos, J.~Neira, R.~Siegwart, and C.~Cadena, ``Empty cities: Image
  inpainting for a dynamic-object-invariant space,'' in \emph{2019
  International Conference on Robotics and Automation (ICRA)}.\hskip 1em plus
  0.5em minus 0.4em\relax IEEE, 2019, pp. 5460--5466.

\bibitem{gupta2018social}
A.~Gupta, J.~Johnson, L.~Fei-Fei, S.~Savarese, and A.~Alahi, ``Social gan:
  Socially acceptable trajectories with generative adversarial networks,'' in
  \emph{IEEE Conference on Computer Vision and Pattern Recognition (CVPR)}, no.
  CONF, 2018.

\bibitem{marchetti2020multiple}
F.~Marchetti, F.~Becattini, L.~Seidenari, and A.~Del~Bimbo, ``Multiple
  trajectory prediction of moving agents with memory augmented networks,''
  \emph{IEEE Transactions on Pattern Analysis and Machine Intelligence}, 2020.

\bibitem{kalman1960new}
R.~E. Kalman, ``A new approach to linear filtering and prediction problems,''
  1960.

\end{thebibliography}

\end{document}